# Common human diseases prediction using machine learning based on survey data

**Jabir Al Nahian[1], Abu Kaisar Mohammad Masum[1], Sheikh Abujar[2], Md. Jueal Mia[1]**
[1]Department of Computer Science and Engineering, Faculty of Science and Information Technology, Daffodil International University, Dhaka, Bangladesh
[2]Department of Computer Science, School of Computer Science and Engineering, Florida Institute of Technology, Florida, United States

**ABSTRACT**

In this era, the moment has arrived to move away from disease as the primary emphasis of medical treatment. Although impressive, the multiple techniques that have been developed to detect the diseases. In this time, there are some types of diseases COVID-19, normal flue, migraine, lung disease, heart disease, kidney disease, diabetics, stomach disease, gastric, bone disease, autism are the very common diseases. In this analysis, we analyze disease symptoms and have done disease predictions based on their symptoms. We studied a range of symptoms and took a survey from people in order to complete the task. Several classification algorithms have been employed to train the model. Furthermore, performance evaluation matrices are used to measure the model's performance. Finally, we discovered that the part classifier surpasses the others.

*Corresponding Author:*

Jabir Al Nahian
Department of Computer Science and Engineering, Faculty of Science and Information Technology
Daffodil International University
Dhaka, Bangladesh
Email: jabir15-10414@diu.edu.bd

## 1. INTRODUCTION

In this modern world, we cannot think our regular life without technology. Artificial intelligence (AI) is one of the core parts of computer science and technology. Creating a sort of AI that's so modern it can itself make AI substances with indeed more prominent intelligence might change man-made innovation until the end of time. Such substances would outperform human insights and reach superhuman accomplishments. Another algorithm within the field of machine learning is data mining which is a fast-growing area. Because it extracts important data from a mountain of datasets and uses it for decision-making tasks, data mining is one of the more well-known concepts in machine learning. Data mining is demonstrated to be an awesome apparatus for investigating modern roads to consequently examine, visualize, and reveal designs in the information that encourage the decision-making process. By immediately providing high-quality diagnosis results and significantly reducing or eliminating the need for human involvement, artificial intelligence, a fast-developing computer technologies in the area of healthcare image identification, has also benefited in sickness prediction. Machine learning and deep learning, two important fields of AI, have lately gained a lot of traction in clinical applications. For diseases diagnosis, systems based on deep learning for support are being created utilizing CT and X-ray samples. Computerized sickness identification has emerged as a crucial area of study in medical advances as a result of rapid population growth. A computerized illness identification framework aids clinicians in disease diagnosis by providing precise, consistent, and quick findings, as well as lowering the death rate [1]. Nowadays, many people are not mindful of their well-being.

There are some people who are not curious about getting to the doctor and that's why people of younger ages are also getting into serious diseases at that time. This type of problem has a lot of complications that change over time. Despite the fact that various strategies have been established, none of them can produce an exact and dependable result. Work with physicians, doctors, and other health experts is a feature of all the modalities. As a result, a system that can function without hospital supplies or personnel may be an effective alternative [2]. It is more necessary to identify consistent risk variables and construct a prediction model for several illnesses than it is to do so for a specific disease. For instance, a patient with hyperlipidemia or Hypertension is more likely to experience cardiovascular problems than someone in good condition. Hyperlipidemia and hypertension are related conditions [3]. Many medical datasets are nowadays easily accessible for research in a variety of medical specialties. Therefore, managing enormous amounts of data by a human is challenging, if not impossible. As a result, computer-based procedures that are more successful are replacing traditional techniques. The use of computers improves accuracy while saving both money and time [4]. Common human diseases have recently had a significant impact on the medical field and the global economy. Scientists, Doctors, and specialists are working on new techniques to diagnose diseases more quickly, such as the creation of autonomous disease detection systems.

Chang *et al.* [5] made a two-phase research approach for predicting hyperlipidemia and hypertension at the same time. They began by selecting specific risk variables for both these two diseases using four data mining methodologies and then utilized the voting principle to discover the shared risk factors. After that, they built multiple predictive models for hyperlipidemia and hypertension using the multivariate adaptive regression splines (MARS) approach. Maleki *et al.* [6] proposed a method for determining the level of lung cancer. In this research, a K-nearest neighbors (KNN) approach is used to diagnose the stage of patients' disease, with a genetic algorithm utilized for effective feature identification to minimize dataset dimensionality and improve classifier speed. The optimal value for k is discovered via an experimental technique to increase the accuracy of the presented algorithm. The proposed method was tested on a lung cancer database and found to be 100 percent accurate. Bang *et al.* [7] developed a multi-classification method based on ML to make a distinction between the gut microbiome and the six diseases listed below: chronic fatigue syndrome/myalgic encephalomyelitis, acquired immune deficiency syndrome, juvenile idiopathic arthritis, multiple sclerosis, colorectal cancer. and stroke. To create the prediction model, they utilized the plentiful microbes at five taxonomic levels as characteristics in only 696 samples obtained from various research. Four multi-class classifiers and two feature selecting approaches, including forwarding selection and backward removal, were used to create classification models. Kunjir *et al.* [8] proposed a method for efficient and advanced disease prediction based on historical training data. Analyzing and evaluating different data methods is the best strategy. For each disease algorithm training data example, the datasets chosen for implementation purposes comprise more than 20 medical relevant attributes. Heart disease, breast cancer, arthritis, and diabetes are among the medical datasets chosen for the research. The Naive Bayes (NB) method was chosen to implement in this project after assessing the prediction accuracy and latency test results. Besag and Newell [9] using an epidemiological dataset of COVID-19 patients from South Korea. Muhammad *et al.* [10] established a model for predicting COVID-19 affected patients' recoveries. To create the models, the decision tree, KNN algorithms, support vector machine (SVM), logistic regression, Naive Bayes, and random forest (RF) were directly implemented on the dataset using the Python programming language presented and demonstrated a Geographical Analysis Machine Learning method for detecting tiny illness clusters. A secondary goal is to go over some frequent difficulties in applying clustering tests to epidemiology data. For the classification of breast cancer disease. Shamrat *et al.* [11] employed some supervised classification approaches. SVM, KNN, RF, decision tree are examples of early breast cancer prediction algorithms. As a result, we used specificity, sensitivity the f1 score, and total accuracy to assess the breast cancer dataset. The results of the breast cancer prediction performance analysis show that SVM had the best results, with a classification accuracy of 97.07 percent. NB and RF, on the other hand, have the second-highest forecast accuracy. Islam *et al.* [12] present a deep learning method that uses a convolutional neural network (CNN) and long short-term memory (LSTM) to diagnose COVID-19 from X-ray images. In this system, deep feature extraction is performed using CNN, and the extracted features are identified by LSTM. The dataset in this system included 1525 COVID-19 pictures from 4575 X-ray scans. The testing results showed that their proposed technique achieved an accuracy of 99.4%, a specificity of 99.2%, an AUC of 99.9%, a sensitivity of 99.3%, and a F1-score of 98.9%. Rajdhan *et al.* [13] proposed study uses data mining techniques such Naive Bayes, decision tree, random forest, and logistic regression to determine the patient's total risk and estimating the likelihood of cardiac disease. Consequently, this study compares the effectiveness of various machine learning techniques. Priya *et al.* [14] analyzed liver patient datasets in order to develop classification algorithms for predicting liver disease. The initial phase involves applying the min-max normalization approach to the actual liver disease datasets that were retrieved from the UCI repository. In the second phase of the liver dataset forecast, PSO feature selection is used to obtain a subset (data) of the standardized whole liver patient datasets that only contains important features. The data set is then subjected to categorization algorithms in the third phase. The accuracy will be calculated in the fourth phase using the root mean error value and root mean square value. Rahman *et al.* [15] used five types of supervised classification algorithms are used in this research, there are logistic regression, SVM, KNN, random forest, and decision tree. The execution of different classification methods was assessed on distinctive estimation procedures such as accuracy, recall, precision, specificity, and f--1 score. In a large community pediatric clinic, Gabrielsen *et al.* [16] developed controls and children who tested negative for autism during the widespread screening. Following the screening, medical evaluations were conducted to ascertain the pattern models (autism, language delay, or typical). Unaware of participants' diagnosis status, licensed psychologists with toddler and autism expertise assessed two 10-minute video samples of participants' autistic evaluations, evaluating five behavioral patterns: responsive, conducting, verbalizing, play, and responding to name. Reviewers were asked to give their opinions on autism referrals based purely on 10-minute assessments. Some of the systems are built using a pre-trained model using transfer learning, while others are implemented using bespoke networks. Machine learning and data science are two more fields that are being utilized to diagnose, prognostic, predict, and forecast disease outbreaks [17]-[27].

After seeing such kind of problem, we got a thought to make a machine to predict human diseases by their symptoms. The goal of this work is to use data-mining techniques to determine common diseases and build a prediction model for these diseases. We have done with eleven diseases and dealt with 53 symptoms under a survey which collects from a vast number of people. By doing this we have collected our data and also work with a different type of algorithm like random forest, logistic regression KNN, and SVM. In our research, we calculated many performance evaluation criteria and compared the results to select the best classifier in the working situation. The part classifier produces the best result in terms of metrics, according to the study of the collected results.

The organization of this paper is listed below: in section 2, the study methodology is described, along with a quick rundown of the dataset, the implementation plan, and the classifier algorithms. The outcome of the experiment and other findings are presented in section 3. Section 4 undertakes a thorough evaluation of related studies to identify any unresolved issues. Section 5 concludes everything in the end.

## 2. METHOD

This part contains the following sections: implementation methodology, data analysis and description, algorithm summary. This section explains how we went about completing this project. Below is a full description of all of the sub-sections.

### 2.1. Implementation procedure

The purpose of this work is to achieve disease prediction. Many important characteristics, especially disease symptoms, are considered to ensure an accurate prediction. Figure 1 shows the numerous processes we followed to finish this project.

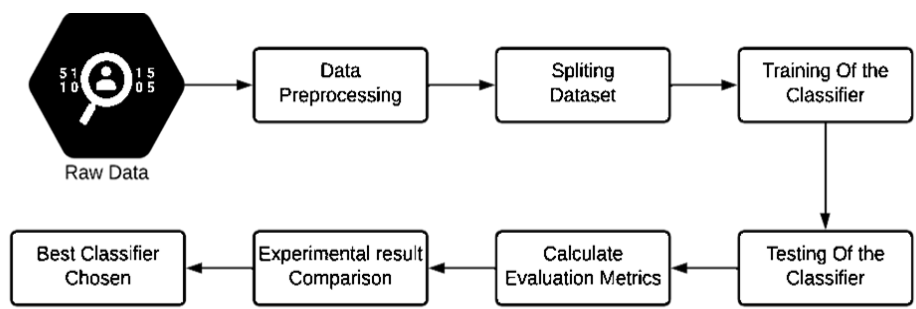

Figure 1. Flowchart of diseases prediction using data mining technique based on symptoms

### 2.1.1. Data collection

First and foremost, we have prepared a 53-question survey based on disease symptoms. Then, we obtained data from a sizable number of respondents utilizing this survey. Then, in order to feed this data into the classifier, we used various preprocessing procedures. To label a specific question, only one variable is used. To identify all of the questions, a total of 53 variables are employed.

### 2.1.2. Data preprocessing

Data preprocessing is the procedure in which input is modified if required. The set of data may contain incomplete characteristics. The median, mean, or other measurements for that property can be used to fill in the blanks. Ultimately, the set of data is randomized to guarantee that the data is distributed evenly.

### 2.1.3. Apply algorithm

Our provided data is divided into the training and testing sets after preprocessing. In this case, 70 percent of the complete data set was used for the training process. The remaining 30% of the entire data set was used for the testing process. This division is done at random. The training process learns the data set from its properties, while the testing process predicts the data set's outcome and assesses predictive accuracy. Following that, the four classifiers, Random Forest, SVM, logistic regression and KNN were trained using the training data. After trained the classifiers, we used testing data to predict the current disease condition.

### 2.1.4. Calculate evaluation

To use these criteria, we found the best classifier to predict in this situation. The formulas below were used to calculate a number of performance indicators in percent using the confusion matrix the classifier produced.

$$\text{Accuracy} = \frac{TP+TN}{TP+FN+FP+TN} \times 100\%$$

$$\text{Sensitivity or Recall or True Positive Rate (TPR)} = \frac{TP}{TP+FN} \times 100\%$$

$$\text{Precision} = \frac{TP}{TP+FP} \times 100\%$$

$$\text{F1 Score} = 2 \times \frac{Precision \times Recall}{Precision+Recall} \times 100\%$$

## 2.2. Data description and analysis

A survey was employed to acquire the information for this research. The questionnaire consisted of 53 questions in total. Both personal and disease symptoms variables are considered in these 53 inquiries. A total of 53 attributes are utilized to categorize all of the queries. The dataset has 52 independent variables and one dependent variable shown in the Table 1. And Table 2 lists all of the variables and their potential values. In order to complete this task, 1443 individual records were needed. 70% of the data is utilized to train the classifier, while 30% is used for testing.

## 2.3. Classifier description

The random forest classification method is a supervised algorithm for learning that divides up random tree groupings into different categories to produce forests. It may be applied to issues with regression and classification. It's a popular method for solving classification difficulties. It picks samples randomly from a particular dataset. It uses data samples to generate decision trees, which are subsequently used to make predictions [28]. Then, using the voting method choose the appropriate solution in the Figure 2. While developing the trees, the random forest contributes more randomness to the pattern. When dividing a node, it searches for the best trait from a specified distribution rather than the most significant characteristic. There is a lot more variability as a result, which produces a better prediction accuracy. Because it is an ensemble learning technique, random forest outperforms a single decision tree. The overfitting problem is reduced by averaging the results.

In reality, the "SVM" is a supervised machine learning method that can resolve regression and classification issues. However, it is primarily employed to address classification issues [29]. Every piece of data is represented as a point in an n-dimensional space (where n is the number of attributes we have), with each feature's value being the score at a particular place in the SVM classifier. SVM is used to choose the most nodes that contribute to the creation of the hyperplane. The algorithm is known as an SVM, and support vectors are also the algorithm's maximum examples. Look at the Figure 3 to see how two distinct groups are classified using a decision hyperplane. Next, we identify the hyper-plane which clearly distinguishes the class labels to complete identification. Simply put, support vectors are also the positions of each accuracy assessment. The classification algorithm is a frontier that separates the two categories (hyper-plane/line) the most effective.

Table 1. Attribute with their possible values

| Variable | Full-form | Variable type | Possible value |
|---|---|---|---|
| Dyhof | Occasional fever | Independent | Yes (1), No (0) |
| Mof | Measure of fever | Independent | 98-102(1), 102-105(2), Normal (3) |
| Amc | Almost cough | Independent | Yes (1), No (0) |
| Toc | Types of cough | Independent | Dry (1), Blood (2), Mucus (3), Normal (4) |
| Ft | Feel tired | Independent | Yes (1), No (0) |
| Vm | Vomiting | Independent | Yes (1), No (0) |
| Tov | Types of vomiting | Independent | Stomach bloated and vomiting (1), Vomiting (2), Nausea (3) |
| Ftotp | Following types of throat problems | Independent | Throat pain (1), Voice change (2), Sore throat (3), None (4) |
| Rsptp | Respiratory problems | Independent | Breath weakness (1), Shortness of breath (2), None (3) |
| Sos | Sense of smell | Independent | No olfactory power (1), Not get smell properly (2), Olfactory power is ok (3) |
| Dodi | Occasional diarrhea | Independent | Yes (1), No (0) |
| Hpyr | Head problems | Independent | Dizziness (1), Severe headache (2), Normal headache (3), None (4) |
| Oryb | Occasional rashes | Independent | Yes (1), No (0) |
| Hrn | Runny nose | Independent | Yes (1), No (0) |
| Flitb | Feel less in the body | Independent | Yes (1), No (0) |
| Sfkd | Suffer in depression | Independent | Yes (1), No (0) |
| Snha | Stiff neck | Independent | Yes (1), No (0) |
| Dyep | Eye's pain | Independent | Yes (1), No (0) |
| Pobsf | Pain on both sides of forehead | Independent | Yes (1), No (0) |
| Ybin | Body numbs | Independent | Yes (1), No (0) |
| Tslt | Tolerate sound, light, touch | Independent | Yes (1), No (0) |
| Ftope | Following types of problems experience | Independent | Chest pain (1), Chest pressure (2), Chest discomfort (3), Chest throbbing (4), Sweating (5), Chest tightening (6), None (7) |
| Thbt | Type heartbeat | Independent | Decreases rapidly (1), Increase rapidly (2), Normal (3) |
| Slc | Lose consciousness | Independent | Yes (1), No (0) |
| Npc | Pain problems experience | Independent | Neck pain (1), Jaw pain (2), Spinal pain (3), None (4) |
| Atw | Ability to work | Independent | Yes (1), No (0) |
| Lwwr | Lose weight without reason | Independent | Yes (1), No (0) |
| Tpof | Types of appetite | Independent | Excessive appetite (1), Appetite depression (2), Normal (3) |
| Urpm | Urination problems | Independent | More (1), Less (2), Not at all (3), normal (4) |
| Obib | Observation in body | Independent | Weight gain (1), Swelling in the body (2), Swelling of eyes and face (3), None (4) |
| Scyh | Sleep condition | Independent | More (1), Less (2), Normal (3), Not at all (4) |
| Abdp | Abdominal pain | Independent | Complete abdominal pain (1), Lower Abdominal pain (2), Flatulence (3), None (4) |
| Diyh | Indigestion | Independent | Yes (1), No (0) |
| Bsdf | Black stools during defecation | Independent | Yes (1), No (0) |
| Asys | Airway sore | Independent | Yes (1), No (0) |
| Dwpd | Drink water per day | Independent | 1-2L (1), 3-5 (2), 5-8(3) |
| Dysbv | Blurred vision | Independent | Yes (1), No (0) |
| Wbdul | Wound body dry up late | Independent | Yes (1), No (0) |
| Ioynf | Infections on your nails or fingers | Independent | Yes (1), No (0) |
| Imhad | Irritable mood | Independent | Yes (1), No (0) |
| Iibd | Itching in body | Independent | Yes (1), No (0) |
| Nacht | Noticed any changes height | Independent | Increase rapidly (1), Decrease (2), Fixed (3) |
| Bppf | Body pain problem | Independent | Pain to sit up (1), Muscle pain (2), pain to bend down (3), Pain in the joints (4), Tingling in the hands and feet (5), None (6) |
| Spcy | Speak clearly | Independent | Yes (1), No (0) |
| Rtcoo | Respond to the call of others | Independent | Yes (1), No (0) |
| Kaeoo | Keep an eye on others | Independent | Yes (1), No (0) |
| Dyew | Empathy | Independent | Yes (1), No (0) |
| Ladyo | like aloneness | Independent | Yes (1), No (0) |
| Guiy | Gesture unusual | Independent | Yes (1), No (0) |
| Syaay | scared | Independent | Yes (1), No (0) |
| Oaay | Overly abusive | Independent | Yes (1), No (0) |
| Rtswaa | Repeat the same word again and again | Independent | Yes (1), No (0) |
| Disease | All disease name | Dependent | Diseases name labelling (a, b, c, d, e, f, g, h, I, j, k, x) |

Table 2. Diseases name their possible values

| Full-form | Possible values | Using label encoder |
|---|---|---|
| Covid 19 | a | 0 |
| Normal flue | b | 1 |
| Migraine | c | 2 |
| Heart disease | d | 3 |
| Lung disease | e | 4 |
| Kidney disease | f | 5 |
| Stomach disease | g | 6 |
| Gastric | h | 7 |
| Diabetics | i | 8 |
| Bone disease | j | 9 |
| Autism | k | 10 |
| No diseases | x | 11 |

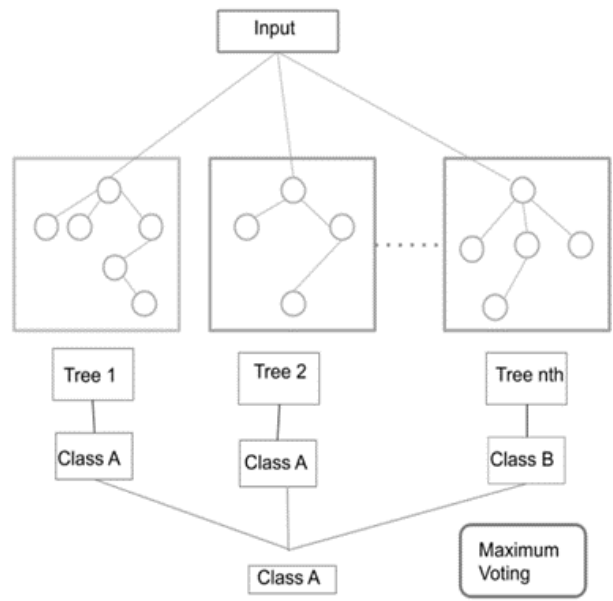

Figure 2. The random forest method is depicted in the diagram

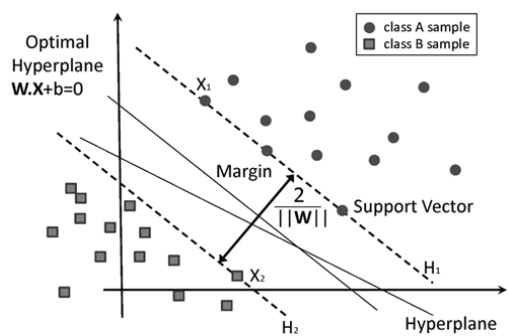

Figure 3. SVM diagram

The supervised learning approach of logistic regression has been used to predict the categorical outcome variable using only a collection of individual variables [30]. This is an important and strong method since it can provide possibilities and identify updated information using those discrete and continuous data. Figure 4 shows that the logistic (Sigmoid) model is a mathematical equation that maps anticipated outcomes to probabilities. It can convert any actual value between 0 and 1 into another. This method's major implication is that the predicted output must be classified and that the input parameter must not be multi-collinear.

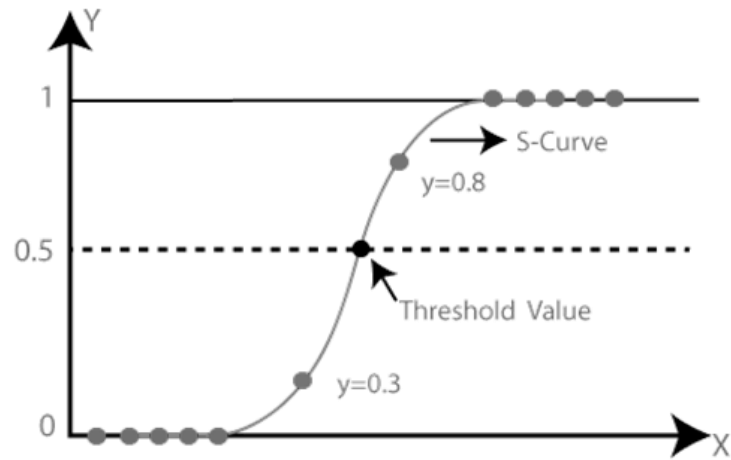

Figure 4. Logistic regression transformation

The KNN method is one of the core principles of learning algorithms. It is predicated on the Supervised Learning approach. The KNN approach is used to address both classification and regression problems. Feature matching is the foundation of the KNN method. A straightforward, understandable, and flexible machine learning technique is KNN.

KNN has uses in a variety of fields, including economics, politics, medical, computer vision, and video recognition. Financial institutions utilize credit ratings to predict a customer's credit rating [31]. The KNN approach allocates the new case to the category that is most similar to the classifications, implying that the current particular instance and previous examples are comparable (look at the Figure 5). KNN represents the number of nearest neighbors. The most important thing to take into account is the quantity of neighbors. When there are two courses, K is often an odd number. The procedure is known as the closest neighbor algorithm when K=1.

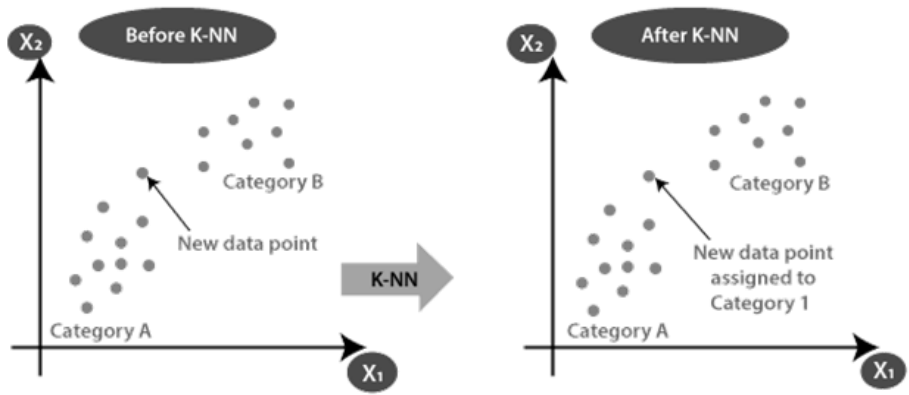

Figure 5. KNN algorithm classification

## 3. RESULT AND DISCUSSION

Since this is essentially a multiclass problem, the classifier produced a 12*12 confusion matrix. Table 3 shows the resulting matrix for each of the classifiers. Accuracy, F1, precision, and recall scores are calculated from the above confusion matrix to evaluate this work. Table 3 shows the results of numerous performance evaluation metrics. Table 3 demonstrates that the part classifier surpasses the other four classifications algorithm when results are examined as a whole. The part classifier has the highest accuracy of all the classifiers for all classes 88.2, 88.1, 88.5, and 88.2. Other Table 3 and Table 4 results also support the part classifier. After applying SVM, random forest, and logistic regression algorithm we found the best classifier algorithm as random forest. The result is also shown below the bar chart Figure 6.

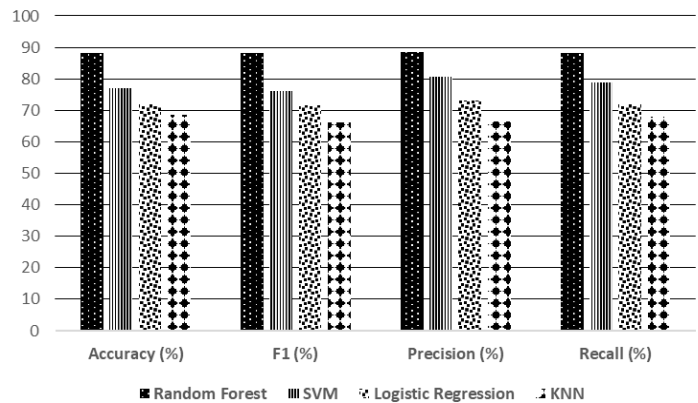

Figure 6. Classifier's performance

Table 3. Comparison of four classifier's performance

| Classifier | Class name | Accuracy (%) | F1 (%) | Precision (%) | Recall (%) |
|---|---|---|---|---|---|
| Random Forest | Covid 19 | 99.3 | 93.6 | 96.4 | 88.0 |
| | Normal flue | 97.6 | 92.1 | 90.6 | 69.4 |
| | Migraine | 96.4 | 81.4 | 94.6 | 81.6 |
| | Heart disease | 99.8 | 97.1 | 98.4 | 93.7 |
| | Lung disease | 99.5 | 87.5 | 100 | 81.3 |
| | Kidney disease | 99.5 | 91.7 | 100 | 94.6 |
| | Stomach disease | 99.5 | 93.3 | 99.8 | 91.4 |
| | Gastric | 100 | 100 | 100 | 100 |
| | Diabetics | 99.3 | 94.1 | 96.0 | 92.3 |
| | Bone disease | 95.5 | 96.7 | 91.9 | 92.7 |
| | Autism | 93.5 | 88.7 | 84.1 | 83.8 |
| | No problem | 92.6 | 88.9 | 81.7 | 79.3 |
| SVM | Covid 19 | 98.8 | 88.9 | 93.4 | 80.0 |
| | Normal flue | 93.8 | 80.3 | 75.7 | 85.5 |
| | Migraine | 96.2 | 81.4 | 94.6 | 71.4 |
| | Heart disease | 100 | 99.0 | 83.3 | 81.4 |
| | Lung disease | 99.5 | 87.5 | 100 | 81.3 |
| | Kidney disease | 98.3 | 93.2 | 95.4 | 91.6 |
| | Stomach disease | 100 | 99.5 | 100 | 100 |
| | Gastric | 99.0 | 91.7 | 88.0 | 86.7 |
| | Diabetics | 98.6 | 87.5 | 84.7 | 80.7 |
| | Bone disease | 89.5 | 83.1 | 81.9 | 79.5 |
| | Autism | 82.7 | 74.8 | 81.8 | 74.7 |
| | No problem | 89.0 | 79.6 | 84.1 | 84.7 |
| Logistic Regression | Covid 19 | 98.1 | 83.3 | 87.0 | 80.0 |
| | Normal flue | 91.1 | 69.9 | 70.5 | 69.4 |
| | Migraine | 91.6 | 69.8 | 78.4 | 77.3 |
| | Heart disease | 89.8 | 86.3 | 81.5 | 79.6 |
| | Lung disease | 100 | 96.7 | 98.3 | 94.7 |
| | Kidney disease | 99.0 | 83.3 | 90.9 | 86.9 |
| | Stomach disease | 100 | 100 | 100 | 100 |
| | Gastric | 100 | 100 | 100 | 100 |
| | Diabetics | 97.1 | 87.5 | 81.8 | 79.2 |
| | Bone disease | 91.6 | 84.4 | 78.6 | 75.2 |
| | Autism | 81.5 | 70.3 | 72.3 | 80.5 |
| | No Problem | 86.3 | 80.1 | 84.7 | 86.6 |
| KNN | Covid 19 | 82.1 | 77.9 | 76.7 | 73.5 |
| | Normal flue | 83.9 | 73.8 | 75.6 | 72.9 |
| | Migraine | 89.4 | 72.9 | 84.4 | 73.8 |
| | Heart disease | 89.4 | 83.5 | 76.1 | 73.4 |
| | Lung disease | 87.6 | 82.3 | 79.4 | 77.3 |
| | Kidney disease | 97.1 | 80.0 | 77.1 | 73.8 |
| | Stomach disease | 93.1 | 83.3 | 91.4 | 85.9 |
| | Gastric | 96.4 | 88.2 | 86.7 | 85.3 |
| | Diabetics | 93.3 | 89.8 | 83.5 | 77.6 |
| | Bone disease | 81.5 | 79.7 | 74.3 | 71.1 |
| | Autism | 73.5 | 72.1 | 69.9 | 80.5 |
| | No problem | 80.8 | 79.9 | 76.4 | 73.1 |

Table 4. Overall comparison of four classifier's performance

| Model | Accuracy (%) | F1 (%) | Precision (%) | Recall (%) |
|---|---|---|---|---|
| Random forest | 88.2 | 88.1 | 88.5 | 88.2 |
| SVM | 77.0 | 76.2 | 80.6 | 78.7 |
| Logistic regression | 71.9 | 71.7 | 73.0 | 71.9 |
| KNN | 68.7 | 66.1 | 66.4 | 67.9 |

## 4. COMPARATIVE ANALYSIS

In this analysis, we analysis some resource which related with our work. We got only three papers related to multiple diseases prediction. But find out some lakes and problems in their research. Chang et al. made a two-phase research approach for predicting hyperlipidemia and hypertension at the same time. They began by selecting specific risk variables for both these two diseases using six data mining methodologies and then utilized the voting principle to discover the shared risk factors. After that, they built multiple predictive models for hyperlipidemia and hypertension using the multivariate adaptive regression splines (MARS) approach [5]. Bang et al. developed a multi-classification method based on ML to make a distinction between the gut microbiome from the six diseases. To create the prediction model, they used the abundance of microorganisms at five taxonomic levels as characteristics in only 696 samples obtained from various research [7]. Kunjir et al. proposed a method for efficient and advanced disease prediction based on historical training data. Analyzing and evaluating different data methods is the best strategy. For each disease algorithm training data example, the datasets chosen for implementation purposes comprise more than 20 medical relevant attributes. Heart disease, breast cancer, arthritis, and diabetes are among the medical datasets chosen for the research [8]. Here, we given below theirs work details and their result. But our research is very unique at this time. Because we successfully predict human diseases based on their given symptoms. All the related work functionality is given in Table 5.

Table 5. Disease prediction related work

| Article | Disease name | Functionality | Observed features | Models/algorithms | Results |
|---|---|---|---|---|---|
| [5] | Hypertension and Hyperlipidemia | Using a two-phase analysis approach, they forecast hyperlipidemia and hypertension at the same time | To create a multivariate predictive model for hypertension and hyperlipidemia, they employed the multivariate adaptive regression splines (MARS) approach | Logistic regression CART CHAID MARS | Accuracy rate: 93.07% |
| [7] | Juvenile idiopathic arthritis, colorectal cancer, multiple sclerosis, myalgic encephalomyelitis, chronic fatigue syndrome, and acquired immune deficiency syndrome | A few microbes' abundances have been used as a flag to forecast a variety of diseases. They anticipated in this work that utilizing a multi-classification ML technique, they could discriminate the gut microbiota from six illnesses | They used four multi-class classifiers and different feature selection strategies, using forward choice and backward removal, to create classification techniques | KNN LMT SVM LogitBoost | Accuracy rate: 83.1% |
| [8] | Diabetes, breast cancer, heart dataset | Create a simple decision-making system that can identify and extract previously unknown patterns, connections, and theories linked to numerous diseases from previous database files of various illnesses. | The developed scheme can answer challenging questions for recognizing a specific condition and may also help healthcare professionals make patient care decisions that existing decision support systems couldn't | Naive Base J48 | Accuracy rate: Diabetes dataset: 76.30% Breast cancer dataset: 71.45% Heart dataset: 83.49% |

## 5. CONCLUSION AND FUTURE WORK

This task mostly consists of predicting an individual's symptoms and identifying the disease. The many data mining techniques used to achieve this result. The major goal of this research is to integrate data mining and machine learning approaches to provide credible results for common human diseases. A total of 70% and 30% of data is utilized to train and test the classifier, respectively, to complete this task. We examined numerous quality assessment criteria to evaluate the effective classification algorithm. We found that part classifier outperforms every other data mining technique. In the future, we'll work with bigger datasets with more attributes and employ more data mining techniques.

# BIOGRAPHIES OF AUTHORS

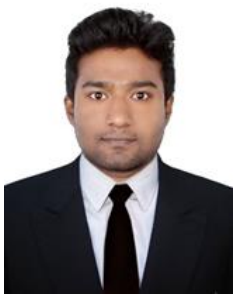
**Jabir Al Nahian** received the Bachelor's degree in Computer Science and Engineering from Daffodil International University, Bangladesh. He is currently Researcher in the Computational Intelligence Lab, Bangladesh specializing in providing machine learning solutions for expertise. His current research interests lie in the area of data science, natural language processing, machine learning, deep learning and particularly in areas pertaining to their application for the Bangla language. He is an active researcher and reviewer several international conferences and journals. He can be contacted at email: jabir15-10414@diu.edu.bd and jabirnahian009@gmail.com.

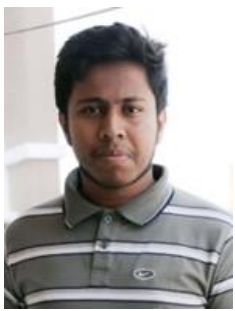
**Abu Kaisar Mohammad Masum** completed his B.Sc. from Daffodil International University (DIU) in Bangladesh. Now he is a Lecturer in the Dept. of CSE, DIU. Previously he worked as Research Assistant (RA) in Apurba-DIU Research & Development Lab. He has a number of Scopus indexed publications in international and national journals and conference proceedings. Broadly, his methodological research focuses on the Application of Machine Learning, Natural Language Processing (NLP), and Data mining. He currently works on different areas of NLP and Adaptive algorithms. Mr. Masum has received Best Researcher Award-2021 organized by DIU and awarded as Best Performed Faculty Member of DIU. He also received 'In Recognition of Scholar Publication in Reputed Indexed Journal' award for the year of 2019 by DIU. Mr. Masum is a Supervisor of the DIU - NLP and Machine Learning Research LAB. He can be contacted at email: abu.cse@diu.edu.bd.

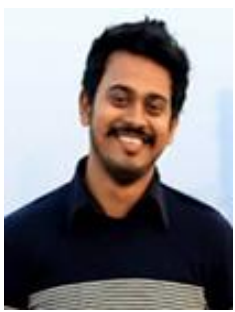
**Seikh Abujar** has obtained his M.Sc. degree in Computer Science from Jahangirnagar University, Bangladesh. Previously he completed his B.Sc. degree in Computer Science and Engineering at Daffodil International University (DIU), Bangladesh. Currently pursuing his PhD in Computer Science at Florida Institute of Technology in Melbourne, Florida. At present he is working as a Lecturer in the department of Computer Science and Engineering in Independent University, Bangladesh (IUB). Prior to joining IUB, he was a faculty member at DIU from September 2017 to January 2021 and at Britannia University (BU) from February 2015 to August 2017. He has a number of articles published in several Scopus Indexed proceedings of IEEE, Springer and Elsevier, and has served on several international conferences and journals as a reviewer. Broadly, his methodological research focuses on application of machine learning and data mining. he currently works on different areas of natural language processing and adaptive algorithms. Many of Mr. Abujar's and his students' research work have won "Best research paper - award" in many International Conferences. During his career at DIU, he was Founder and Supervisor of the DIU - NLP and Machine Learning Research LAB. He can be contacted at email: sabujar2021@my.fit.edu.

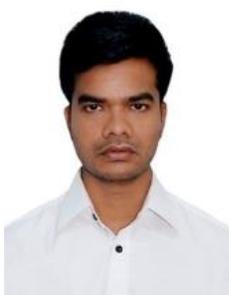
**Md. Jueal Mia** received his B.Sc. (Honors) and M.Sc. in Computer Science and Engineering from Jahangirnagar University, Dhaka, Bangladesh in 2014 and 2015 respectively. Currently, he works as a Senior Lecturer in the Department of Computer Science and Engineering, Daffodil International University, Dhaka, Bangladesh. He has published many articles in international journals and conference proceedings. His research interest is mainly on artificial intelligence, computer vision, machine learning, expert systems, data mining, and natural language processing. He can be contacted at email: mjueal02@gmail.com.